%% file: main.tex
\documentclass[11pt,a4paper]{article}
\usepackage[hyperref]{emnlp2020}
\usepackage{times}
\usepackage{latexsym}

\usepackage{multirow}
\usepackage{paralist}
\usepackage{color, colortbl}
\usepackage{enumitem}
\usepackage[normalem]{ulem}

\usepackage{microtype}
\usepackage{graphicx}
\usepackage{subcaption}
\usepackage{adjustbox}
\usepackage{wrapfig}
\usepackage{array}
\usepackage{tabu}
\usepackage{makecell}
\usepackage{tabulary}
\usepackage{booktabs}
\usepackage{wrapfig}

\aclfinalcopy 
\def\aclpaperid{1049} 

\input{math_commands.tex}

\newcommand{\TOTTO}{\textsc{ToTTo}}
\newcommand{\pagetitle}{\textrm{page-title}}
\newcommand{\sectiontitle}{\textrm{section-title}}
\newcommand{\sectiontext}{\textrm{section-text}}
\newcommand{\delete}{\textrm{deletion}}
\newcommand{\ambiguity}{\textrm{decontext}}
\newcommand{\highlight}{\textrm{highlight}}
\newcommand{\final}{\textrm{final}}
\newcommand{\Dtrain}{\mD_{\textrm{orig-train}}}
\newcommand{\Dskewtrain}{\mD_{\textrm{train}}}
\newcommand{\Ddev}{\mD_{\textrm{dev}}}
\newcommand{\Dtest}{\mD_{\textrm{test}}}
\newcommand{\Dtestoverlap}{\mD_{\textrm{test-overlap}}}
\newcommand{\Dtestnonoverlap}{\mD_{\textrm{test-nonoverlap}}}
\newcommand{\Ddevoverlap}{\mD_{\textrm{dev-overlap}}}
\newcommand{\Ddevnonoverlap}{\mD_{\textrm{dev-nonoverlap}}}

\definecolor{canaryyellow}{rgb}{1.0, 0.94, 0.0}
\definecolor{bananayellow}{rgb}{1.0, 0.88, 0.21}

\title{ToTTo: A Controlled Table-To-Text Generation Dataset}

\author{Ankur P. Parikh$^\spadesuit$ \ \ \ Xuezhi Wang$^\spadesuit$ \ \ \ Sebastian Gehrmann$^\spadesuit$ \\\ \textbf{Manaal Faruqui}$^\spadesuit$ \ \ \  \textbf{Bhuwan Dhingra}$^{\clubsuit}$\thanks{\ \ Work done during an internship at Google.} \ \ \ \textbf{Diyi Yang}$^{\spadesuit \diamondsuit}$ \ \ \ \textbf{Dipanjan Das}$^\spadesuit$ \\
$^\spadesuit$ Google Research, New York, NY \\
$^\diamondsuit$ Georgia Tech, Atlanta, GA \\
$^\clubsuit$ Carnegie Mellon University, Pittsburgh, PA \\ \\
\texttt{totto@google.com}
}

\date{}

\begin{document}
\maketitle
\begin{abstract}
We present \TOTTO, an open-domain English table-to-text dataset with over 120,000 training examples that proposes a controlled generation task: given a Wikipedia table and a set of highlighted table cells, produce a one-sentence description. To obtain generated targets that are natural but also faithful to the source table, 
we introduce a dataset construction process where annotators directly revise existing candidate sentences from Wikipedia. We present systematic analyses of our dataset and annotation process as well as 
results achieved by several state-of-the-art baselines. While usually fluent, existing methods often hallucinate phrases that are not supported by the table, suggesting that this dataset can serve as a useful research benchmark for high-precision conditional text generation.\footnote{\TOTTO \, is available at \url{https://github.com/google-research-datasets/totto}.}
\end{abstract}

\section{Introduction}

\begin{table*}[t]
\centering
\footnotesize
\begin{tabular}{c|c|c|c|c|c}
\multicolumn{6}{l}{\textbf{Table Title}: Gabriele Becker}   \\
\multicolumn{6}{l}{\textbf{Section Title}: International Competitions} \\ 
\multicolumn{6}{l}{\textbf{Table Description}: None}   \\ \toprule
\textbf{Year}                                  & \textbf{Competition}                                          & \textbf{Venue}                         & \textbf{Position} & \textbf{Event}                        & \textbf{Notes}         \\ \midrule
\multicolumn{6}{l}{\textbf{Representing Germany}}                                                                                                                                                                                            \\ \hline
1992                                           & World Junior Championships                                    & Seoul, South Korea                     & 10th (semis)      & 100 m                                 & 11.83                  \\ \hline
                                               &                                                               &                                        & 7th               & 100 m                                 & 11.74                  \\ \cline{4-6}
\multirow{-2}{*}{1993}                         & \multirow{-2}{*}{European Junior Championships}               & \multirow{-2}{*}{San Sebasti\'an, Spain} & 3rd               & 4x100 m relay                         & 44.60                   \\ \hline
                                               &                                                               &                                        & 12th (semis)      & 100 m                                 & 11.66 \scriptsize(wind: +1.3 m/s)   \\ \cline{4-6}
\multirow{-2}{*}{1994}                         & \multirow{-2}{*}{World Junior Championships}                  & \multirow{-2}{*}{Lisbon, Portugal}     & 2nd               & 4x100 m relay                         & 44.78                  \\ \hline
\cellcolor[HTML]{FFEF00}                       & \cellcolor[HTML]{FFEF00}                                      &                                        & 7th (q-finals)    & \cellcolor[HTML]{FFEF00}100 m         & 11.54                  \\ \cline{4-6}
\multirow{-2}{*}{\cellcolor[HTML]{FFEF00}1995} & \multirow{-2}{*}{\cellcolor[HTML]{FFEF00}World Championships} & \multirow{-2}{*}{Gothenburg, Sweden}   & 3rd               & \cellcolor[HTML]{FFEF00}4x100 m relay & 43.01                  \\ \bottomrule
\multicolumn{6}{l}{\begin{tabular}[c]{@{}l@{}}\textbf{Original Text}: After winning the German under-23 100 m title, she was selected to run at the 1995 World Championships \\ in Athletics both individually and in the relay.\end{tabular}}            \\ %
\multicolumn{6}{l}{\textbf{Text after Deletion}: she at the 1995 World Championships in both individually and in the relay.} \\ 
\multicolumn{6}{l}{\begin{tabular}[c]{@{}l@{}}\textbf{Text After Decontextualization}: Gabriele Becker competed at the 1995 World Championships \\ in both individually and in the relay. \end{tabular}}  \\ 
\multicolumn{6}{l}{\textbf{Final Text}: Gabriele Becker competed at the 1995 World Championships both individually and in the relay.}     \\ 
\bottomrule
\end{tabular}
\caption{Example in the \TOTTO \, dataset. The goal of the task is given the table, table metadata (such as the title), and set of highlighted cells, to produce the final text. Our data annotation process revolves around annotators iteratively revising the original text to produce the final text. }
\label{tab:example}
\end{table*}

Data-to-text generation~\citep{kukich1983design,mckeown1992text} is the task of generating a target textual description $\vy$ conditioned on source content $\vx$ in the form of structured data such as a table.
Examples include generating sentences given biographical data~\citep{lebret2016neural},  textual descriptions of restaurants given meaning representations~\citep{novikova2017e2e}, basketball game summaries given boxscore statistics~\citep{wiseman2017challenges}, and generating fun facts from superlative tables in Wikipedia~\citep{flip2019}.

Existing data-to-text tasks have provided an important test-bed for neural generation models~\citep{sutskever2014sequence,bahdanau2014neural}. Neural models are known to be prone to \emph{hallucination}, i.e., generating text that is fluent but not faithful to the source~\citep{vinyals2015neural,koehn2017six,lee2018hallucinations,tian2019sticking} and it is often easier to assess faithfulness of the generated text when the source content is structured~\citep{wiseman2017challenges,dhingra2019handling}. Moreover, structured data can also test a model's ability for reasoning and numerical inference~\citep{wiseman2017challenges} and for building representations of structured objects~\citep{liu2017table}, providing an interesting complement to tasks that test these aspects in the NLU setting~\citep{pasupat2015compositional,chen2019tabfact,dua2019drop}.

However, constructing a data-to-text dataset can be challenging on two axes: task design and annotation process. First, tasks with open-ended output like summarization~\cite{mani1999advances,lebret2016neural,wiseman2017challenges} lack explicit signals for models on what to generate, which can lead to subjective content and evaluation challenges~\citep{kryscinski2019neural}.
On the other hand, data-to-text tasks that are limited to verbalizing a fully specified meaning representation~\citep{gardent2017webnlg} do not test a model's ability to perform inference and thus remove a considerable amount of challenge from the task.

Secondly, designing an annotation process to obtain natural but also clean targets is a significant challenge. One strategy employed by many datasets is to have annotators write targets from scratch~\citep{banik2013kbgen,wen2015semantically,gardent2017creating} which can often lack variety in terms of structure and style~\citep{gururangan2018annotation,poliak2018hypothesis}. An alternative is to pair naturally occurring text with tables~\citep{lebret2016neural,wiseman2017challenges}. While more diverse, naturally occurring targets are often noisy and contain information that cannot be inferred from the source.  This can make it problematic to disentangle modeling weaknesses from data noise. 

In this work, we propose \TOTTO, an \textit{open-domain} table-to-text generation dataset that introduces a novel task design and annotation process to address the above challenges. First, \TOTTO \ proposes a \textit{controlled} generation task: given a Wikipedia table and a set of highlighted cells as the source $\vx$, the goal is to produce a single sentence description $\vy$. The highlighted cells  identify  portions of potentially large tables that the target sentence should describe, without specifying an explicit meaning representation to verbalize.

For dataset construction, to ensure that targets are natural but also faithful to the source table, we request annotators to \textit{revise} existing Wikipedia candidate sentences into target sentences, instead of asking them to write new target sentences~\citep{wen2015semantically,gardent2017creating}. Table~\ref{tab:example} presents a simple example from \TOTTO \ to illustrate our annotation process. The table and \emph{Original Text} were obtained from Wikipedia using heuristics that collect pairs of tables $\vx$ and sentences $\vy$ that likely have significant semantic overlap.
This method ensures that the target sentences are \textit{natural}, although they may only be partially related to the table. Next, we create a clean and controlled generation task by requesting annotators to highlight a subset of the table that supports the original sentence and revise the latter iteratively to produce a final sentence (see \S\ref{sec:annotation}). 
For instance, in Table~\ref{tab:example}, the annotator has chosen to highlight a set of table cells (in yellow) that support the original text. They then deleted phrases from the original text that are not supported by the table, e.g., \emph{After winning the German under-23 100 m title}, and replaced the pronoun \emph{she} with an entity \emph{Gabriele Becker}. The resulting final sentence (\emph{Final Text}) serves as a more suitable generation target than the original sentence.
This annotation process makes our dataset well suited for \textit{high-precision} conditional text generation.

Due to the varied nature of Wikipedia tables, \TOTTO \ covers a significant variety of domains while containing targets that are completely faithful to the source (see Table~\ref{tab:basketball-example} and the Appendix for more complex examples). Our experiments demonstrate that state-of-the-art neural models struggle to generate faithful results, despite the high quality of the training data.  These results suggest that our dataset could serve as a useful benchmark for controllable data-to-text generation.

\section{Related Work}
\label{sec:motivation}
\TOTTO \, differs from existing datasets in both task design and annotation process as we describe below. A summary is given in Table~\ref{tab:dataset-comparison}.


\begin{table*}[!tb]
\centering
\scriptsize
\setlength\tabcolsep{3pt}
\begin{tabular}{@{}lrllll@{}}
\toprule
 \textbf{Dataset} & \textbf{Train Size} & \textbf{Domain} & \textbf{Target Quality} & \textbf{Target Source} & \textbf{Content Selection}  \\ \hline 
Wikibio~\citep{lebret2016neural} & 583K & Biographies & Noisy & Wikipedia & Not specified \\
Rotowire~\citep{wiseman2017challenges}  & 4.9K & Basketball & Noisy & Rotowire & Not specified \\
WebNLG~\cite{gardent2017webnlg} & 25.3K & 15 DBPedia categories & Clean & Annotator Generated & Fully specified  \\
E2E~\citep{novikova2017e2e} & 50.6K & Restaurants & Clean & Annotator Generated & Partially specified \\
LogicNLG~\citep{chen2020logical} & 28.5K & Wikipedia (open-domain) & Clean & Annotator Generated & Columns via entity linking \\
\textbf{\TOTTO} & \textbf{120K} & \textbf{Wikipedia (open-domain)} & \textbf{Clean} & \textbf{Wikipedia (Annotator Revised)} & \textbf{Annotator highlighted} \\ \bottomrule
\end{tabular}
\caption{Comparison of popular data-to-text datasets. \TOTTO \ combines the advantages of annotator-generated and fully natural text through a revision process.}
\label{tab:dataset-comparison}
\end{table*}

\paragraph{Task Design} 
Most existing table-to-text datasets are restricted in topic and schema such as \textsc{WeatherGov}~\citep{liang2009learning}, \textsc{RoboCup}~\citep{chen2008learning}, Rotowire~\citep[basketball]{wiseman2017challenges}, E2E~\citep[restaurants]{novikova2016crowd,novikova2017e2e}, KBGen~\citep[biology]{banik2013kbgen}, and Wikibio~\citep[biographies]{lebret2016neural}. In contrast, \TOTTO \, contains tables with various schema spanning various topical categories all over Wikipedia.
Moreover, \TOTTO \, takes a different view of content selection compared to existing datasets. Prior to the advent of neural approaches, generation systems typically separated content selection (\emph{what to say}) from surface realization (\emph{how to say it})~\citep{reiter1997building}. Thus many generation datasets only focused on the latter stage~\citep{wen2015semantically,gardent2017webnlg}.
However, this decreases the task complexity, since neural systems have already been quite powerful at producing fluent text. 
Some recent datasets~\citep{wiseman2017challenges,lebret2016neural} have proposed incorporating content selection into the task by framing it as a summarization problem. However, summarization is much more subjective, which can make the task underconstrained and difficult to evaluate~\citep{kryscinski2019neural}.
We place \TOTTO\ as a middle-ground where the highlighted cells provide some guidance on the topic of the target but still leave a considerable amount of content planning to be done by the model.

\paragraph{Annotation Process} There are various existing strategies to create the reference target $\vy$. One strategy employed by many datasets is to have annotators write targets from scratch given a representation of the source~\citep{banik2013kbgen,wen2015semantically,gardent2017creating}. While this will result in a target that is faithful to the source data, it often lacks variety in terms of structure and style~\citep{gururangan2018annotation,poliak2018hypothesis}. Domain-specific strategies such as presenting an annotator an image instead of the raw data~\citep{novikova2016crowd}~are not practical for some of the complex tables that we consider.
Other datasets have taken the opposite approach: finding real sentences on the web that are heuristically selected in a way that they discuss the source content~\citep{lebret2016neural,wiseman2017challenges}. This strategy typically leads to targets that are natural and diverse, but they may be noisy and contain information that cannot be inferred from the source~\citep{dhingra2019handling}.
To construct \TOTTO, we ask annotators to revise existing candidate sentences from Wikipedia so that they only contain information that is supported by the table. This enables \TOTTO \ to maintain the varied language and structure found in natural sentences while producing cleaner targets. The technique of editing exemplar sentences  has been used in semiparametric generation models~\citep{guu2018generating,pandey2018exemplar,peng2019text} and crowd-sourcing small, iterative changes to text has been shown to lead to higher-quality data and a more robust annotation process~\citep{little2010turkit}. \citet{perez2018bootstrapping} also employed a revision strategy to construct a cleaner evaluation set for Wikibio~\cite{lebret2016neural}. 

 Concurrent to this work,~\citet{chen2020logical} proposed LogicNLG which also uses Wikipedia tables, although omitting some of the more complex structured ones included in our dataset. Their target sentences are annotator-generated and their task is significantly more uncontrolled due to the lack of annotator highlighted cells.
 
\section{Preliminaries}
\label{sec:preliminaries}
Our tables come from English Wikipedia articles and thus may not be regular grids.\footnote{In Wikipedia, some cells may span multiple rows and columns.  See Table 1 for an example.}  
For simplicity, we define a table $\vt$ as a set of cells $\vt = \{\vc_j\}_{j=1}^{\tau}$ where $\tau$ is the number of cells in the table. Each cell contains: \textbf{(1)} a string value, \textbf{(2)} whether or not it is a row or column header, \textbf{(3)} the row and column position of this cell in the table, \textbf{(4)} The number of rows and columns this cell spans.



Let $\vm = (m_{\pagetitle}, m_{\sectiontitle}, m_{\sectiontext})$ indicate table metadata, i.e, the page title, section title, and up to the first 2 sentences of the section text (if present) respectively. These fields can help provide context to the table's contents.
Let $\vs = (s_1,...,s_{\eta})$ be a sentence of length $\eta$. 
We define an annotation example\footnote{An annotation example is different than a task example since the annotator could perform a different task than the model.} 
$\vd = (\vt, \vm, \vs)$ a tuple of table, table metadata, and sentence. Here, $\mD = \{ \vd_n \}_{n=1}^{N}$ refers to a dataset of annotation examples of size $N$. 

\section{Dataset Collection}\label{sec:raw-annotations}
We first describe how to obtain annotation examples $\vd$ for subsequent annotation. To prevent any overlap with the Wikibio dataset~\citep{lebret2016neural}, we do not use infobox tables. We employed three heuristics to collect tables and sentences:
\paragraph{Number matching} We 
search for tables and sentences on the same Wikipedia page that overlap with a non-date number of at least 3 non-zero digits. This approach captures most of the table-sentence pairs that describe statistics (e.g., sports, election, census, science, weather).

\paragraph{Cell matching} We 
extract a sentence 
if it has tokens matching at least 3 distinct cell contents from the \textbf{same row} in the table.
The intuition is that most tables are structured, and a row is usually used to describe a complete event.

\paragraph{Hyperlinks} The above heuristics only consider sentences and tables on the same page. We also find examples where a sentence $\vs$ contains a hyperlink to a page with a title that starts with \emph{List} (these pages typically only consist of a large table). If the table $\vt$ on that page also has a hyperlink to the page containing $\vs$, then we consider this to be an annotation example. Such examples typically result in more diverse examples than the other two heuristics, but also add more noise, since the sentence may only be distantly related to the table.

Using the above heuristics we obtain a set of examples $\mD$. 
We then sample a random subset of tables for annotation, excluding tables with formatting issues: 191,693 examples for training, 11,406 examples for development, and 11,406 examples for test. 
Among these examples, 35.8\% were derived from number matching, 29.4\% from cell matching, and 34.7\% from hyperlinks. 


\begin{table*}[!tbh]
\scriptsize
\begin{tabular}{p{3.6cm} p{3.6cm} p{3.5cm} p{3.5cm} }
\toprule
\textbf{Original} & \textbf{After Deletion} & \textbf{After Decontextualization} & \textbf{Final} \\ \midrule
He later raced a Nissan Pulsar and then a Mazda 626 in this series, with a highlight of finishing runner up to Phil Morriss in the 1994 Australian Production Car Championship. & He {\color{red} \sout{later}} raced a Nissan Pulsar and then a Mazda 626 {\color{red} \sout{in this series, with a highlight of}} finishing runner up {\color{red} \sout{to Phil Morriss}} in the 1994 Australian Production Car Championship. & \noindent{\color{blue} \underline{Murray Carter}} raced a Nissan Pulsar and finished as a runner up in the 1994 Australian Production Car Championship. & Murray Carter raced a Nissan Pulsar and finished as runner up in the 1994 Australian Production Car Championship.  \\ \hline
On July 6, 2008, Webb failed to qualify for the Beijing Olympics in the 1500 m after finishing 5th in the US Olympic Trials in Eugene, Oregon with a time of 3:41.62.
& On July 6, 2008, Webb {\color{red} \sout{failed to qualify for the Beijing Olympics in the 1500 m after}} finishing 5th in the {\color{red} \sout{US}} Olympic Trials in Eugene, Oregon with a time of 3:41.62.
& On July 6, 2008, Webb finishing 5th in the Olympic Trials in Eugene, Oregon with a time of 3:41.62.
& On July 6, 2008, Webb {\color{orange} finished} 5th in the Olympic Trials in Eugene, Oregon, with a time of 3:41.62.  \\ \hline
Out of the 17,219 inhabitants, 77 percent were 20 years of age or older and 23 percent were under the age of 20. & \noindent{\color{red} \sout{Out of the} } 17,219 inhabitants { \color{red} \sout{, 77 percent were 20 years of age or older and 23 percent were under the age of 20}}. & \noindent{\color{blue} \underline{Rawdat Al Khail} } had a population of  17,219 inhabitants. & Rawdat Al Khail had a population of 17,219 inhabitants. \\ \bottomrule
\end{tabular}
\caption{Examples of annotation process. Deletions are indicated in red strikeouts, while added named entities are indicated in underlined blue. Significant grammar fixes are denoted in orange.}
\label{tab:annotation-examples}
\end{table*}

\section{Data Annotation Process}
\label{sec:annotation}

The collected annotation examples are noisy since a sentence $\vs$ may only be partially supported by the table $\vt$. 
We thus define an annotation process that guides annotators through incremental changes to the original sentence. 
This allows us to measure annotator agreement at every step of the process, which is atypical in existing generation datasets.




The primary annotation task consists of the following steps: \textbf{(1)} Table Readability, \textbf{(2)} Cell highlighting, \textbf{(3)} Phrase Deletion, \textbf{(4)} Decontextualization. After these steps we employ a final secondary annotation task for grammar correction. Each of these are described below and more examples are provided in the Table~\ref{tab:annotation-examples}.

\paragraph{Table Readability} If a table is not readable, then the following steps will not need to be completed. This step is only intended to remove fringe cases where the table is poorly formatted or otherwise not understandable (e.g., in a different language). 99.5\% of tables are determined to be readable.

\paragraph{Cell Highlighting}  An annotator is instructed to highlight cells that support the sentence. A phrase is supported by the table if it is either directly stated in the cell contents or meta-data, or can be logically inferred by them. Row and column headers do not need to be highlighted. If the table does not support any part of the sentence, then no cell is marked and no other step needs to be completed. 69.7\% of examples are supported by the table. For instance, in Table~\ref{tab:example}, the annotator highlighted cells that support the phrases \emph{1995}, \emph{World Championships}, \emph{individually}, and \emph{relay}. The set of highlighted cells are denoted as a subset of the table: $\vt_{\highlight} \in \vt$.



\paragraph{Phrase Deletion} This step removes phrases in the sentence unsupported by the selected table cells. Annotators are restricted such that they are only able to delete phrases, transforming the original sentence: $\vs \rightarrow \vs_{\delete}$.
In Table~\ref{tab:example}, the annotator transforms $\vs$ by removing several phrases such as \emph{After winning the German under-23 100 m title}.

On average,  $\vs_{\delete}$ is different from $\vs$ for 85.3\% of examples and while $\vs$ has an average length of 26.6 tokens, this is reduced to 15.9 for $\vs_{\delete}$. We found that the phrases annotators often disagreed on corresponded to verbs purportedly supported by the table. 



\paragraph{Decontextualization} A given sentence $\vs$ may contain pronominal references or other phrases that depend on context.
We thus instruct annotators to identify the main topic of the sentence; if it is a pronoun or other ambiguous phrase, we ask them to replace it with a named entity from the table or metadata. To discourage excessive modification, they are instructed to make at most one replacement.\footnote{Based on manual examination of a subset of 100 examples, all of them could be decontextualized with only one replacement. Allowing annotators to make multiple replacements led to excessive  clarification.} 
This transforms the sentence yet again: $\vs_{\delete} \rightarrow \vs_{\ambiguity}$. In Table~\ref{tab:example}, the annotator replaced \emph{she} with \emph{Gabriele Becker}.

Since the previous steps can lead to ungrammatical sentences, annotators are also instructed to fix the grammar to improve the fluency of the sentence.  We find that $\vs_{\ambiguity}$ is different than $\vs_{\delete}$ 68.3\% of the time, and the average sentence length increases to 17.2 tokens for $\vs_{\ambiguity}$ compared to 15.9 for $\vs_{\delete}$.

\paragraph{Secondary Annotation Task}

Due to the complexity of the task,  $\vs_{\ambiguity}$ may still have grammatical errors, even if annotators were instructed to fix grammar. Thus, a second set of annotators were asked to further correct the sentence and were shown the table with highlighted cells as additional context. This results in the final sentence $\vs_{\final}$. On average, annotators edited the sentence 27.0\% of the time, and the sentence length slightly increased to 17.4 tokens from 17.2. 

\section{Dataset Analysis} 

\begin{table*}[]
\centering
\scriptsize
\begin{tabular}{c|c|c|c|c|c|c|c|c|c|c|c}
 \multicolumn{12}{l}{\textbf{Table Title}: Robert Craig (American football)}   \\
 \multicolumn{12}{l}{\textbf{Section Title}: National Football League statistics} \\ 
 \multicolumn{12}{l}{\textbf{Table Description}:None}  \\
\toprule
\multicolumn{1}{l}{} & \multicolumn{6}{c|}{\textbf{RUSHING}} & \multicolumn{5}{c}{\textbf{RECEIVING}}                                   \\ \hline
\textbf{YEAR}        & \textbf{TEAM} & \textbf{ATT}  & \textbf{YDS}  & \textbf{AVG} & \textbf{LNG} & \textbf{TD} & \textbf{NO.} & \textbf{YDS}  & \textbf{AVG} & \textbf{LNG} & \textbf{TD} \\ \hline
\cellcolor{canaryyellow}1983                 & SF            & 176           & 725           & 4.1          & 71           & 8           & 48           & 427           & 8.9          & 23           & 4           \\ \hline
\cellcolor{canaryyellow}1984                 & SF            & 155           & 649           & 4.2          & 28           & 4           & 71           & 675           & 9.5          & 64           & 3           \\ \hline
\cellcolor{canaryyellow} 1985                 & SF            & 214           & 1050          & 4.9          & 62           & 9           & 92           & 1016          & 11           & 73           & 6           \\ \hline
\cellcolor{canaryyellow} 1986                 & SF            & 204           & 830           & 4.1          & 25           & 7           & 81           & 624           & 7.7          & 48           & 0           \\ \hline
\cellcolor{canaryyellow} 1987                 & SF            & 215           & 815           & 3.8          & 25           & 3           & 66           & 492           & 7.5          & 35           & 1           \\ \hline
\cellcolor{canaryyellow} 1988                 & SF            & 310           & 1502          & 4.8          & 46           & 9           & 76           & 534           & 7.0            & 22           & 1           \\ \hline
\cellcolor{canaryyellow} 1989                 & SF            & 271           & 1054          & 3.9          & 27           & 6           & 49           & 473           & 9.7          & 44           & 1           \\ \hline
\cellcolor{canaryyellow} 1990                 & SF            & 141           & 439           & 3.1          & 26           & 1           & 25           & 201           & 8.0            & 31           & 0           \\ \hline
\cellcolor{canaryyellow} 1991                 & RAI           & 162           & 590           & 3.6          & 15           & 1           & 17           & 136           & 8.0            & 20           & 0           \\ \hline
\cellcolor{canaryyellow} 1992                 & MIN           & 105           & 416           & 4.0            & 21           & 4           & 22           & 164           & 7.5          & 22           & 0           \\ \hline
\cellcolor{canaryyellow} 1993                 & MIN           & 38            & 119           & 3.1          & 11           & 1           & 19           & 169           & 8.9          & 31           & 1           \\ \hline
\textbf{Totals}      & \textbf{-}    & \textbf{1991} & \cellcolor{canaryyellow} \textbf{8189} & \textbf{4.1} & \textbf{71}  & \textbf{56} & \cellcolor{canaryyellow} \textbf{566} & \cellcolor{canaryyellow} \textbf{4911} & \textbf{8.7} & \textbf{73}  & \textbf{17} \\
\bottomrule
 \multicolumn{12}{p{12cm}}{\textbf{Target Text}: Craig finished his eleven NFL seasons with 8,189 rushing yards and 566 receptions for 4,911 receiving yards.}   \\
 \bottomrule
\end{tabular}
 \caption{An example in the \TOTTO \, dataset that involves numerical reasoning over the table structure.}
 \label{tab:basketball-example}
\end{table*}

\label{sec:analysis}
Basic statistics of \TOTTO \ are described in Table~\ref{tab:dataset-stats}. The number of unique tables and vocabulary size attests to the open domain nature of our dataset. Furthermore, while the median table is actually quite large (87 cells),  the median number of highlighted cells is significantly smaller (3). This indicates the importance of the cell highlighting feature of our dataset toward a well-defined text generation task.

\begin{table}[!tb]
\centering
\small
\begin{tabular}{@{}lr@{}}
\toprule
\textbf{Property} & \textbf{Value}  \\ \midrule 
Training set size & 120,761 \\
Number of target tokens  & 1,268,268 \\
Avg Target Length (tokens) & 17.4  \\
Target vocabulary size & 136,777 \\
Unique Tables & 83,141 \\
Rows per table (Median/Avg) & 16 / 32.7 \\
Cells per table (Median/Avg) & 87 / 206.6 \\
No. of Highlighted Cell (Median/Avg) & 3 / 3.55 \\
\midrule
Development set size & 7,700 \\
Test set size & 7,700 \\\bottomrule
\end{tabular}
\caption{\TOTTO \, dataset statistics.}
\label{tab:dataset-stats}
\end{table}


\begin{table}[tbp]
\small
\begin{tabular}{@{}llr@{}}
\toprule
\textbf{Annotation Stage}          & \textbf{Measure}                            & \textbf{Result} \\ \midrule
Table Readability         & Agreement / $\kappa$ & 99.38 / 0.646         \\
Cell Highlighting         & Agreement / $\kappa$ & 73.74 / 0.856          \\
After Deletion            & BLEU-4    & 82.19                      \\
After Decontextualization & BLEU-4    & 72.56                      \\
Final                     & BLEU-4    & 68.98                      \\ \bottomrule
\end{tabular}
\caption{Annotator agreement over the development set. If possible, we measure the total agreement (in \%) and the Fleiss' Kappa ($\kappa$). Otherwise, we report the BLEU-4 between annotators.}
\label{tab:annotator_agreement}
\end{table}

\subsection{Annotator Agreement}
Table~\ref{tab:annotator_agreement} shows annotator agreement 
over the development set for each step of the annotation process. We compute annotator agreement and Fleiss' kappa \cite{fleiss_kappa} for table readability and highlighted cells, and BLEU-4 score between annotated sentences in different stages. 

As one can see, the table readability task has an agreement of 99.38\%.  The cell highlighting task is more challenging. 73.74\% of the time all three annotators completely agree on the set of cells which means that they chose the exact same set of cells. The Fleiss' kappa is $0.856$, which is regarded as ``almost perfect agreement'' ($0.81$ - $1.00$) according to \cite{kappa_table}.

With respect to the sentence revision tasks, we see that the agreement slightly degrades as more steps are performed. We compute single reference BLEU among all pairs of annotators for examples in our development set (which only contains examples where both annotators chose $\vt_{\highlight} \neq \emptyset$). As the sequence of revisions are performed, the annotator agreement gradually decreases in terms of BLEU-4: $82.19 \rightarrow 72.56 \rightarrow 68.98$. This is considerably higher than the BLEU-4 between the original sentence $\vs$ and $\vs_{\final}$ (43.17). 

\subsection{Topics and Linguistic Phenomena}

\begin{figure}[!t]
\centering
\includegraphics[width=1.0\columnwidth]{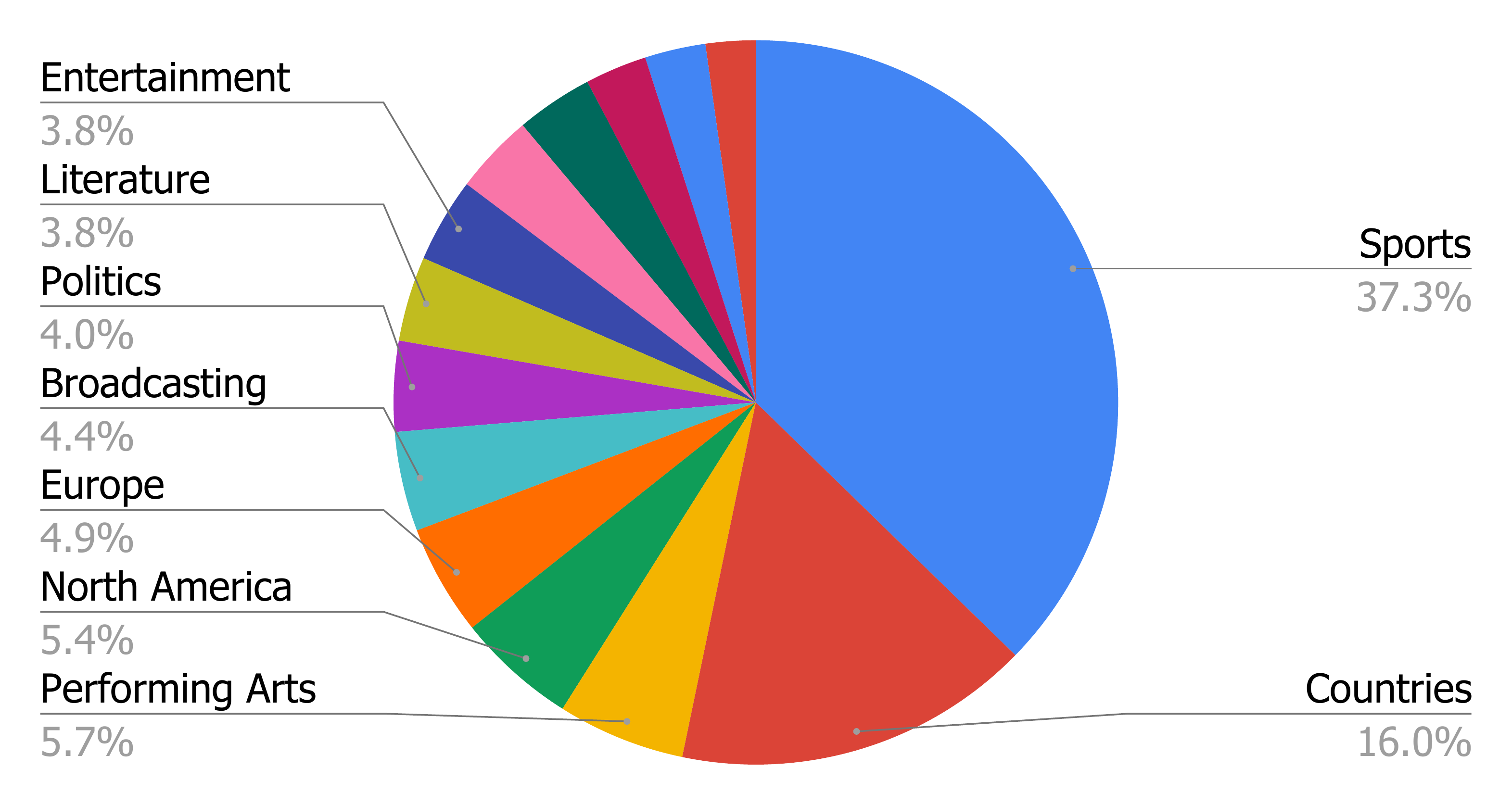}
\caption{Topic distribution of our dataset.}
\label{fig:topics}
\end{figure}

We use the Wikimedia Foundation's topic categorization model~\citep{asthana2018few} to sort the categories of Wikipedia articles where the tables come from into a 44-category ontology.\footnote{\url{https://en.wikipedia.org/wiki/Wikipedia:WikiProject_Council/Directory}}
Figure~\ref{fig:topics} presents an aggregated topic analysis of our dataset.
We found that the \textsl{Sports} and \textsl{Countries} topics together comprise 53.4\% of our dataset, but the other 46.6\% is composed of broader topics such as \textsl{Performing Arts}, \textsl{Politics}, and \textsl{North America}. Our dataset is limited to topics that are present in Wikipedia.

Table~\ref{tab:challenge-types} summarizes the fraction of examples that require reference to the metadata, as well as some of the challenging linguistic phenomena in the dataset that potentially pose new challenges to current systems. Table~\ref{tab:basketball-example} gives one example that requires reasoning (refer to the Appendix for more examples).

\begin{table}[!tbp]
\setlength\tabcolsep{3pt}
\small
\begin{tabular}{@{}lr@{}}
\toprule
\textbf{Types} & \textbf{Percentage} \\ \midrule
Require reference to page title & 82\%\\
Require reference to section title & 19\%\\
Require reference to table description & 3\%\\
Reasoning (logical, numerical, temporal etc.) & 21\%\\
Comparison across rows / columns / cells & 13\%\\
Require background information & 12\%\\
\bottomrule
\end{tabular}
\caption{Distribution of different linguistic phenomena among 100 randomly chosen sentences.}
\label{tab:challenge-types}
\end{table}

\subsection{Training, Development, and Test Splits}
After the annotation process, we only consider examples where the sentence is related to the table, i.e., $\vt_{\highlight} \neq \emptyset$. This initially results in a training set $\Dtrain$ of size 131,849 that we further filter as described below. Each example in the development and test sets was annotated by three annotators. Since the machine learning task uses $\vt_{\highlight}$ as an input, it is challenging to use three different sets of highlighted cells in evaluation. Thus, we only use a single randomly chosen $\vt_{\highlight}$ while using the three $\vs_{\final}$ as references for evaluation. We only use examples where at least 2 of the 3 annotators chose $\vt_{\highlight} \neq \emptyset$, resulting in development and test sets of size 7,700 each.



\paragraph{Overlap and Non-Overlap Sets} Without any modification $\Dtrain$, $\Ddev$, and $\Dtest$ may contain many similar tables. Thus, to increase the generalization challenge, we filter $\Dtrain$ to remove some examples based on overlap with $\Ddev, \Dtest$.

For a given example $\vd$, let $\vh(\vd)$ denote its set of header values and similarly let $\vh(\mD)$ be the set of header values for a given dataset $\mD$. We remove examples $\vd$ from the training set where $\vh(\vd)$ is both rare in the data as well as occurs in either the development or test sets.
Specifically, $\Dskewtrain$ is defined as:
\begin{align*}
\Dskewtrain := \{ \vd : \vh(\vd) \notin \left ( \vh(\Ddev) \cup \vh(\Dtest) \right ) \textrm {or } \\ \textrm{count} \left ( \vh(\vd), \Dtrain \right ) > \alpha  \}.
\end{align*}
The $\textrm{count}(\vh(\vd), \Dtrain)$ function returns the number of examples in $\Dtrain$ with header $\vh(\vd)$. To choose the hyperparameter $\alpha$ we first split the test set as follows:
\begin{align*}
\Dtestoverlap := \{ \vd : \vh(\vd) \in \vh(\Dskewtrain) \} \\
\Dtestnonoverlap := \{ \vd : \vh(\vd) \notin \vh(\Dskewtrain) \} 
\end{align*}
The development set is analogously divided into $\Ddevoverlap$ and $\Ddevnonoverlap$. We then choose $\alpha=5$ so that $\Dtestoverlap$ and $\Dtestnonoverlap$ have similar size. After filtering, the size of $\Dskewtrain$ is 120,761, and $\Ddevoverlap$, $\Ddevnonoverlap$, $\Dtestoverlap$, and $\Dtestnonoverlap$ have sizes $3784$, $3916$, $3853$, and $3847$ respectively.

 \section{Machine Learning Task Construction}
 \label{sec:tasks}

In this work, we focus on the following task:  Given a table $\vt$, related metadata $\vm$ (page title, section title, table section text) and a set of highlighted cells $\vt_{\highlight}$, produce the final sentence $\vs_{\final}$.
Mathematically this can be described as learning a function $\vf : \vx \rightarrow \vy$ where $\vx = (\vt, \vm, \vt_{\highlight})$ and $\vy  = \vs_{\final}$. This task is different from what the annotators perform, since they are provided a starting sentence requiring revision. Therefore, the task is more challenging, as the model must generate a new sentence instead of revising an existing one.

\section{Experiments}
\label{sec:experiments}

\begin{table*}[]
\scriptsize
\centering
\begin{tabular}{@{}lrrrrrr@{}}
\toprule
 \multirow{2}{*}{\raisebox{-\heavyrulewidth}{Model}} & \multicolumn{2}{c}{Overall} & \multicolumn{2}{c}{Overlap Subset} & \multicolumn{2}{c}{Nonoverlap Subset} \\ \cmidrule{2-7} & \textbf{BLEU} & \textbf{PARENT} & \textbf{BLEU} & \textbf{PARENT} & \textbf{BLEU} & \textbf{PARENT} \\ \midrule
BERT-to-BERT (Books+Wiki) & \textbf{44.0} & \textbf{52.6}   & \textbf{52.7} & \textbf{58.4}   & \textbf{35.1} & \textbf{46.8}   \\
BERT-to-BERT (Books)      & 43.9 & \textbf{52.6}   &\textbf{ 52.7} & \textbf{58.4}   & 34.8 & 46.7   \\
Pointer-Generator         & 41.6 & 51.6   & 50.6 & 58.0   & 32.2 & 45.2   \\
\citet{puduppully2019data}          & 19.2    & 29.2      & 24.5    & 32.5      & 13.9   & 25.8      \\ \bottomrule
\end{tabular}
\vspace{-0.05in}
	\caption{Performance compared to multiple references on the test set for the subtable input format with metadata.}
	\label{tab:primary-single-results-test}
\end{table*}

\begin{table*}[!tbp]
\scriptsize
\begin{tabular}{llcccc}
\toprule
 & \textbf{Model} & \multicolumn{1}{l}{\textbf{Fluency (\%)}} & \multicolumn{1}{l}{\textbf{Faithfulness (\%)}} & \multicolumn{1}{l}{\textbf{Covered Cells (\%)}} & \multicolumn{1}{l}{\textbf{Less/Neutral/More Coverage w.r.t. Ref}} \\ \midrule
\multirow{3}{*}{Overall} & \textit{Oracle} & \textit{99.3} & \textit{93.6} & \textit{94.8} & \textit{18.3 / 61.7 / 20.0} \\
 & BERT-to-BERT (Books) & 88.1 & 76.2 & 89.0 & 49.2 / 36.2 / 14.5 \\
 & BERT-to-BERT (Books+Wiki) & 87.3 & 73.6 & 87.3 & 53.9 / 32.9 / 13.2 \\ \hline
\multirow{3}{*}{Overlap} & \textit{Oracle} & \textit{99.6} & \textit{96.5} & \textit{95.5} & \textit{19.8 / 62.8 / 17.4} \\
 & BERT-to-BERT (Books) & 89.6 & 78.7 & 92.1 & 42.0 / 43.7 / 14.3 \\
 & BERT-to-BERT (Books+Wiki) & 89.8 & 81.1 & 91.0 & 47.8 / 39.2 / 13.1 \\ \hline
\multirow{3}{*}{Non-overlap} & \textit{Oracle} & \textit{99.1} & \textit{91.4} & \textit{94.3} & \textit{17.0 / 60.9 / 22.1} \\
 & BERT-to-BERT (Books) & 86.9 & 74.2 & 86.4 & 55.5 / 29.8 / 14.7 \\
 & BERT-to-BERT (Books+Wiki) & 84.8 & 66.6 & 83.8 & 60.1 / 26.6 / 13.3 \\ \bottomrule
\end{tabular}
\vspace{-0.05in}
\caption{Human evaluation over references (to compute \textit{Oracle}) and model outputs. For Fluency, we report the percentage of outputs that were completely fluent. In the last column $X / Y / Z$ means X\% and Z\% of the candidates were deemed to be less and more informative than the reference respectively and Y\% were neutral.}
\label{tab:human-eval}
\end{table*}

\begin{table}[]
\scriptsize
\centering
\begin{tabular}{@{}lrr@{}}
\toprule
\textbf{Data Format} & \textbf{BLEU} & \textbf{PARENT} \\ \midrule
subtable w/ metadata   & 43.9  & 52.6 \\
subtable w/o metadata  & 36.9  & 42.6 \\
full table w/ metadata  & 26.8 & 30.7 \\
full table w/o metadata & 20.9 & 22.2 \\ \bottomrule
\end{tabular}
\caption{Multi-reference performance of different input representations for BERT-to-BERT Books model.}
	\label{tab:input-representations}
\vspace{-0.1in}
\end{table}

We present baseline results on \TOTTO \ by examining three existing state-of-the-art approaches (Note that since our tables do not have a fixed schema it is difficult to design a template baseline).
\vspace{-\topsep}
\begin{itemize}
    \setlength{\parskip}{0pt}
    \setlength{\itemsep}{0pt plus 1pt}
    \item 
    BERT-to-BERT~\citep{rothe2019leveraging}: A Transformer encoder-decoder model~\citep{vaswani2017attention} where the encoder and decoder are both initialized with BERT~\citep{devlin2018bert}. 
    The original BERT model is pre-trained with both Wikipedia and the Books corpus~\citep{zhu2015aligning}, the former of which contains our (unrevised) test targets. 
    Thus, we also pre-train a version of BERT on the Books corpus only, which we consider a more correct baseline. However, empirically we find that both models perform similarly in practice (Table~\ref{tab:primary-single-results-test}).
    \item 
    Pointer-Generator~\citep{see2017get}: A Seq2Seq model with attention and copy mechanism. While originally designed for summarization it is commonly used in data-to-text as well~\citep{gehrmann2018end}.
    \item 
    \citet{puduppully2019data}: A Seq2Seq model with an explicit content selection and planning mechanism designed for data-to-text.
\end{itemize}
Details about hyperparameter settings are provided in the Appendix. Moreover, we explore different strategies of representing the source content
that resemble standard linearization approaches in the literature~\citep{lebret2016neural,wiseman2017challenges}
\vspace{-\topsep}
\begin{itemize}
    \setlength{\parskip}{0pt}
    \setlength{\itemsep}{0pt plus 1pt}
    \item \textbf{Full Table} The simplest approach is simply to use the entire table as the source, adding special tokens to mark which cells have been highlighted.    However, many tables can be very large and this strategy performs poorly. 
    \item \textbf{Subtable} Another option is to only use the highlighted cells $\vt_{highlight} \in \vt$ with the heuristically extracted row and column header for each highlighted cell. This makes it easier for the model to only focus on relevant content but limits the ability to perform reasoning in the context of the table structure (see Table~\ref{tab:decoder_output}). Overall though, we find this representation leads to higher performance.
\end{itemize}
In all cases, the cells are linearized with row and column separator tokens. We also experiment with prepending the table metadata to the source table.\footnote{The table section text is ignored, since it is usually missing or irrelevant.} 

\paragraph{Evaluation metrics} The model output is evaluated using two automatic metrics: BLEU~\cite{papineni2002bleu} and PARENT~\citep{dhingra2019handling}. PARENT is a metric recently proposed specifically for data-to-text evaluation that takes the table into account.  We modify it to make it suitable for our dataset, described in the Appendix. Human evaluation is described in \S~\ref{subsec:humaneval}.



\subsection{Results}

Table~\ref{tab:primary-single-results-test} shows our results against multiple references with the subtable input format. Both the BERT-to-BERT models perform the best, followed by the pointer generator model.\footnote{Note the BLEU scores are relatively high due to the fact that our task is more controlled than other text generation tasks and that we have multiple references.} We see that for all models the performance on the non-overlap set is significantly lower than that of the overlap set, indicating that slice of our data poses significant challenges for machine learning models. We also observe that the baseline that separates content selection and planning performs quite poorly. We attest this to the fact that it is engineered to the Rotowire data format and schema. 


Table~\ref{tab:input-representations} explores the effects of the various input representations (subtable vs. full table) on the BERT-to-BERT model. We see that the full table format performs poorly even if it is the most knowledge-preserving representation. 

\begin{table*}[!h]
\scriptsize
\begin{tabular}{|c|p{3.6cm}|p{3.4cm}|p{3.5cm}|p{3.5cm}|}
\hline 

\textbf{ID}& \textbf{Reference} & \multicolumn{2}{c|}{\textbf{Decoder output (w/ metadata)}} & \textbf{w/o metadata}\\
\hline 
 && \textbf{Full table}  & \textbf{Subtable} &  \textbf{Subtable} \\
\hline \hline
1 & in the 1939 currie cup, western province lost to transvaal by 17–6 in cape town.
& northern transvaal and western province {\color{red}were drawn by 16 in 1989 and in 1995}, western province were defeated by the {\color{red}sharks in durban}.
& the {\color{red}first} currie cup was played in 1939 in {\color{red}transvaal1} at newlands, with western province {\color{red}winning} 17–6.
& the {\color{red}first} {\color{red}scottish} cup was played in 1939 in {\color{red}transvaal1}, where transvaal1 defeated western province 17–6 in the final at newlands in cape town. \\ 
\hline
2 & a second generation of microdrive was announced by ibm in 2000 with increased capacities at 512 mb and 1 gb.
& the microdrive models {\color{red}formed} 512 megabyte and 1 gigabyte in 2000.
& there were {\color{red}512 microdrive models} in 2000: 1 gigabyte.
& \noindent{\color{red}cortete's production} was 512 megabyte.\\
\hline
3 & the 1956 grand prix motorcycle racing season consisted of six grand prix races in five classes: 500cc, 350cc, 250cc, 125cc and sidecars 500cc.
& the {\color{red}1966} grand prix motorcycle racing season consisted of {\color{red}seven} grand prix races in five classes: 500cc, 350cc, 250cc, 125cc and sidecars 500cc.
& the 1956 grand prix motorcycle racing season consisted of {\color{red}eight} grand prix races in five classes: 500cc, 350cc, 250cc, 125cc and sidecars 500cc.
& the {\color{red}1955} grand prix motorcycle racing season consisted of {\color{red}eight} grand prix races in five classes: 500cc, 350cc, 250cc, 125cc and sidecars 500cc. \\
\hline
4 & in travis kelce's {\color{blue}last} collegiate season, he set personal {\color{blue}career highs} in receptions (45), receiving yards (722), yards per receptions (16.0) and receiving touchdowns (8).
&  during the {\color{red}2011} season, travis kelceum {\color{red}caught 76 receptions for 1,612 yards and 14 touchdowns}.
& travis kelce finished the 2012 season with 45 receptions for 722 yards (16.0 avg.) and eight touchdowns.
& kelce finished the 2012 season with 45 catches for 722 yards (16.0 avg.) and eight touchdowns. \\
\hline
5 & in the 2012 film pizza bagel, michael pillarella portrays tommy.
& in 2012, {\color{red}groff} played the role of tommy in the film pizza bagel.
& in 2012, pillarella appeared as tommy in the film pizza bagel.
& \noindent{\color{red} harris} played the role of tommy in the 2012 film pizza bagel.\\
\hline
6 & the album shari addison placed at no. 176 on the billboard 200 along with no. 5 on the gospel albums.
& shari addison's {\color{red}" 5"}, reached number 176 on the billboard 200.
& shari addison charted at number 176 on the {\color{red}us chart} and at number 5 on the {\color{red}us billboard 200}.
& the shari addison peaked at number 176 on the billboard 200 chart. \\
\hline
\end{tabular}
\caption{Decoder output examples from BERT-to-BERT Books models on the development set. The ``subtable with metadata'' model achieves the highest BLEU. Red indicates model errors and blue denotes interesting reference language not in the model output.} 
\label{tab:decoder_output}
\end{table*}

\subsection{Human evaluation} 

\label{subsec:humaneval}
For each of the 2 top performing models in Table~\ref{tab:primary-single-results-test}, we take 500 random outputs and perform human evaluation using the following axes:
\vspace{-\topsep}
\begin{itemize}
    \setlength{\parskip}{0pt}
    \setlength{\itemsep}{0pt plus 1pt}
    \item 
    \textbf{Fluency} - A candidate sentence is fluent if it is grammatical and natural. The three choices are \textit{Fluent}, \textit{Mostly Fluent}, \textit{Not Fluent}.
    \item
    \textbf{Faithfulness} (Precision) - A candidate sentence is considered faithful if all pieces of information are supported by either the table or one of the references. Any piece of unsupported information makes the candidate unfaithful.
    \item 
    \textbf{Covered Cells} (Recall) - Percentage of highlighted cells the candidate sentence covers.
    \item
    \textbf{Coverage with Respect to Reference} (Recall) - We ask whether the candidate is strictly more or less informative than each reference (or neither, which is referred to as neutral).
\end{itemize}

We further compute an oracle upper-bound by treating one of the references as a candidate and evaluating it compared to the table and other references. The results, shown in Table~\ref{tab:human-eval}, attest to the high quality of our human annotations since the oracle consistently achieves high performance. All the axes demonstrate that there is a considerable gap between the model and oracle performance.

This difference is most easily revealed in the last column when annotators are asked to directly compare the candidate and reference. As expected, the oracle has similar coverage to the reference (61.7\% neutral) but both baselines demonstrate considerably less coverage. According to an independent-sample t-test, this difference is significant at a $p<0.001$ level for both baselines. 
Furthermore, the baselines are considerably less faithful than the reference. The faithfulness of both models is significantly lower than the reference ($\chi^2$ test with $p<0.001$). The models do not differ significantly from each other, except for faithfulness in the non-overlap case, where we see a moderate effect favoring the book model. 

\section{Model Errors and Challenges}
\label{sec:errors}

Table~\ref{tab:decoder_output} shows predictions from the BERT-to-BERT Books model to illustrate challenges existing models face.

\paragraph{Hallucination} The model sometimes outputs phrases such as \emph{first}, \emph{winning} that seem reasonable but are not faithful to the table. 
This hallucination phenomenon has been widely observed in other existing data-to-text datasets~\citep{lebret2016neural,wiseman2017challenges}. However, the noisy references in these datasets make it difficult to disentangle model incapability from data noise. Our dataset serves as strong evidence that even when the reference targets are faithful to the source, neural models still struggle with faithfulness. 
\paragraph{Rare topics}
Another challenge revealed by the open domain nature of our task is rare or complex topics at the tail of the topic distribution (Figure~\ref{fig:topics}). For instance, example 2 of Table~\ref{tab:decoder_output} concerns microdrive capacities which is challenging. 
\paragraph{Diverse table structure and numerical reasoning} In example 3, inferring \emph{six} and \emph{five} correctly requires counting table rows and columns. Similarly, in example 4, the phrases \emph{last} and \emph{career highs} can be deduced from the table structure and with comparisons over the columns. However, the model is unable to make these inferences from the simplistic source representation that we used. 
\paragraph{Evaluation metrics} Many of the above issues are difficult to capture with metrics like BLEU since the reference and prediction may only differ by a word but largely differ in terms of semantic meaning. This urges for better metrics possibly built on learned models~\citep{wiseman2017challenges,ma2019results,sellam2020bleurt}. Thus, while we have a task leaderboard, it should not be interpreted as the definitive measure of model performance.

\section{Conclusion}
We presented \TOTTO, a table-to-text dataset that presents a controlled generation task and a data annotation process based on iterative sentence revision. We also provided several state-of-the-art baselines, and demonstrated \TOTTO\ could serve as a useful research benchmark for model and metric development. \TOTTO \, is available at \url{https://github.com/google-research-datasets/totto}.

\section*{Acknowledgements}
The authors wish to thank Ming-Wei Chang, Jonathan H. Clark, Kenton Lee, and Jennimaria Palomaki for their insightful discussions and support. Many thanks also to Ashwin Kakarla and his team for help with the annotations.

\bibliography{main}
\bibliographystyle{acl_natbib}

\appendix

\section{Appendix}

\begin{figure*}[!h]
\centering
\includegraphics[width=1.8\columnwidth]{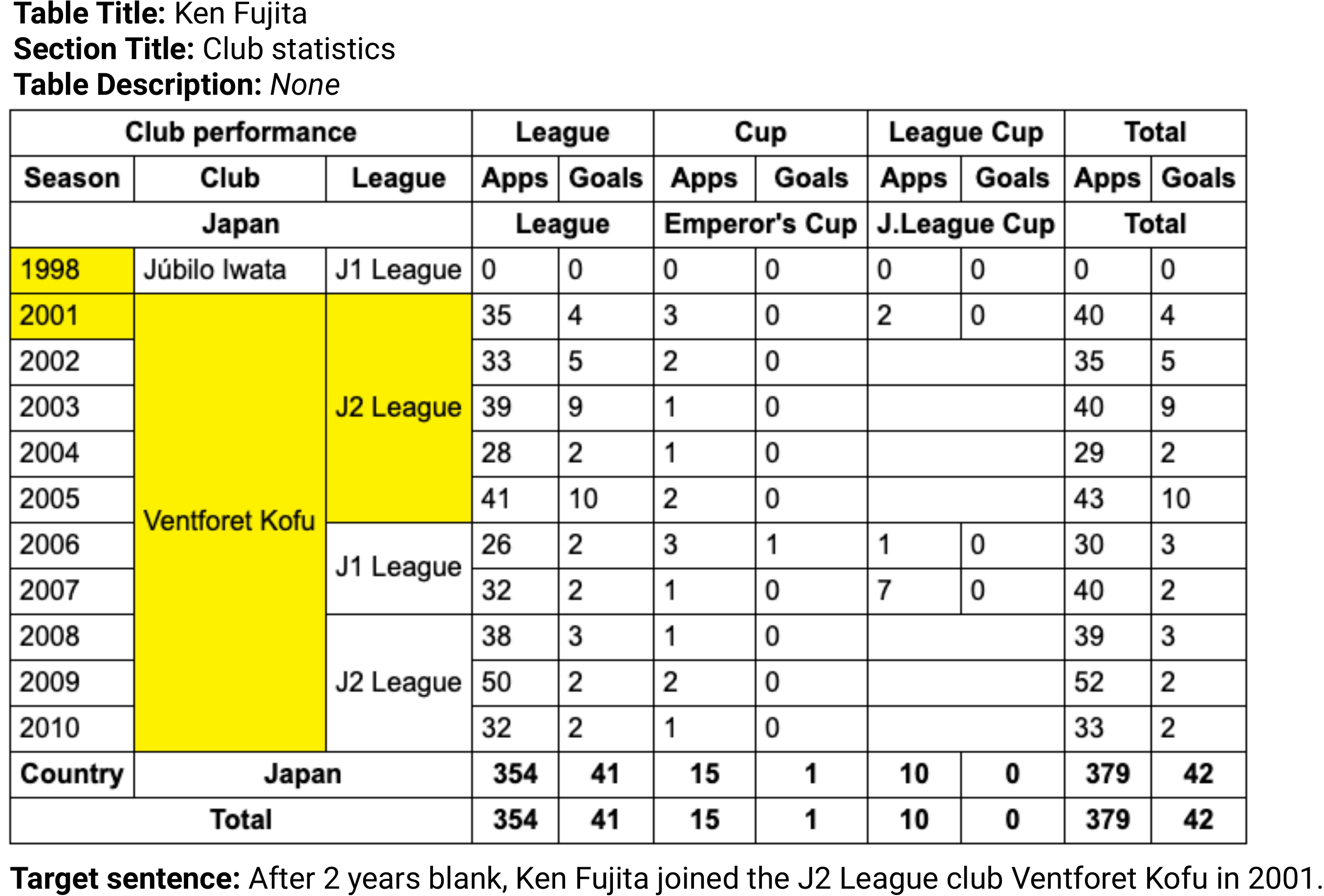}
\caption{\TOTTO \, example with complex table structure and temporal reasoning.}
\label{fig:appendix-ex2}
\end{figure*}

\begin{figure*}[!h]
\centering
\includegraphics[width=1.8\columnwidth]{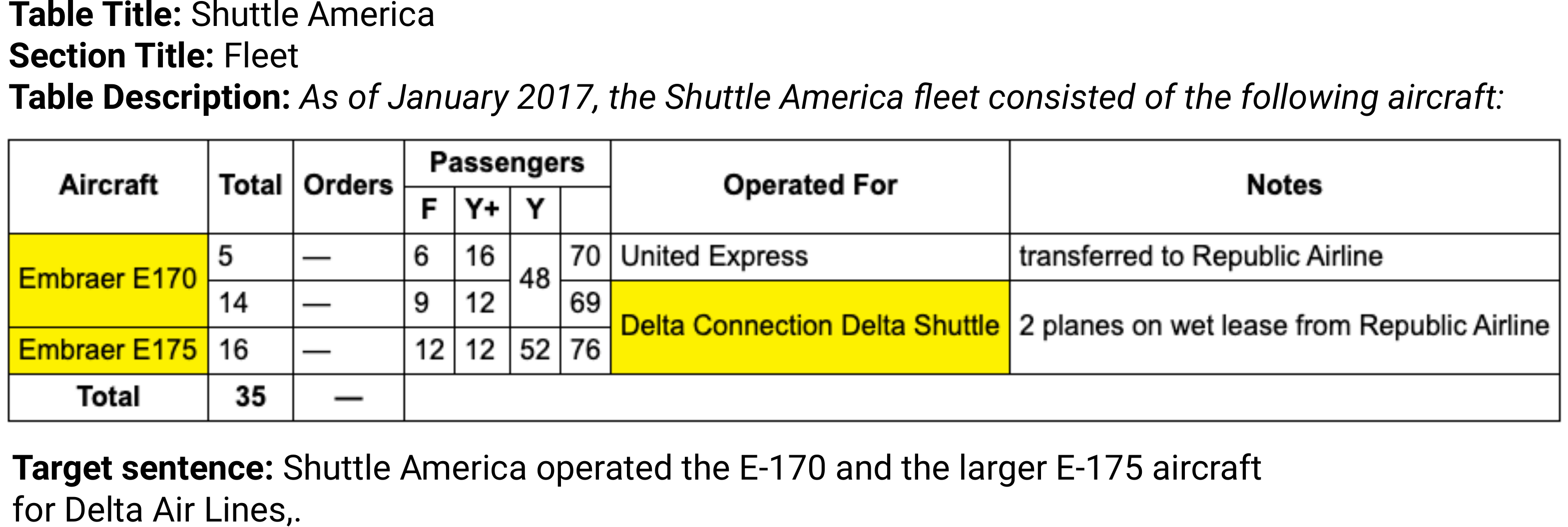}
\caption{\TOTTO \, example with rare topics and complex table structure.}
\label{fig:appendix-ex3}
\end{figure*}

\begin{figure*}[!h]
\centering
\includegraphics[width=1.9\columnwidth]{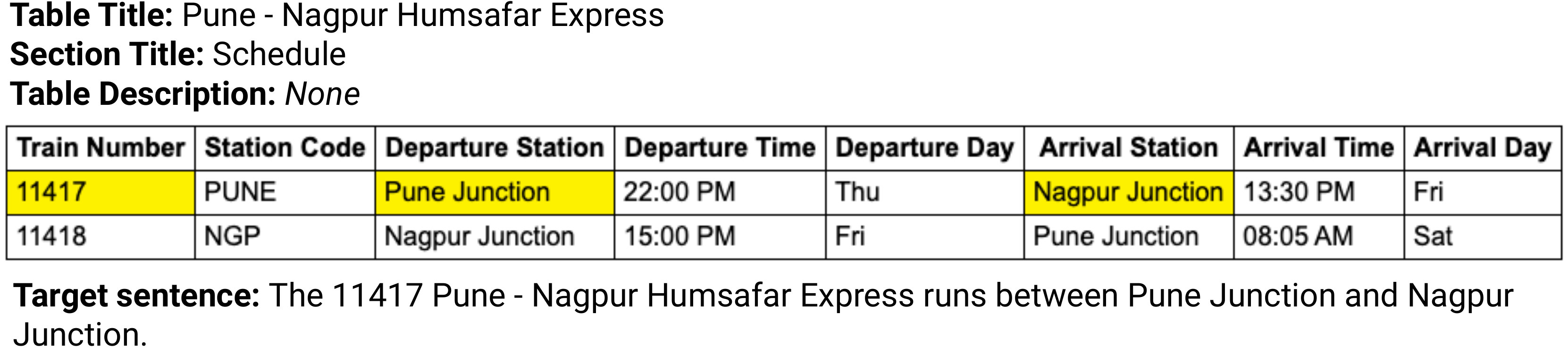}
\caption{\TOTTO \, example with rare topic.}
\label{fig:appendix-ex4}
\end{figure*}

\begin{figure*}[!h]
\centering
\includegraphics[width=1.9\columnwidth]{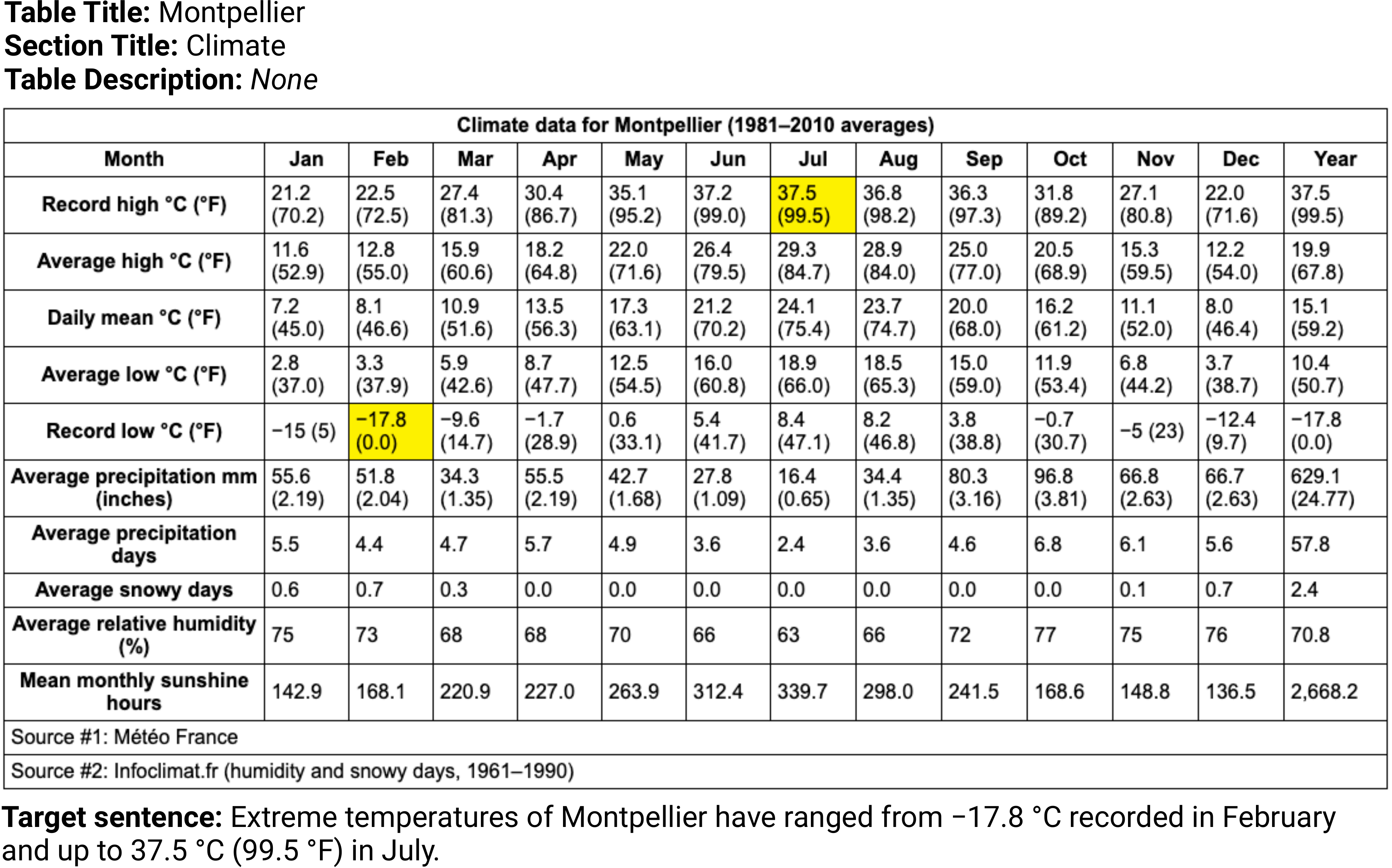}
\caption{\TOTTO \, example with interesting reference language.}
\label{fig:appendix-ex5}
\end{figure*}

The Appendix contains the following contents:
\begin{itemize}
    \item Information about the variant of the PARENT metric~\citep{dhingra2019handling} used for evaluation.
    \item More details about the baselines.
    \item Examples of more complex tables in our dataset (Figure~\ref{fig:appendix-ex2}-Figure~\ref{fig:appendix-ex5}).
\end{itemize}

\subsection{PARENT metric}

PARENT~\citep{dhingra2019handling} is a metric recently proposed  specifically for data-to-text evaluation that takes the table into account.  We modify it to make it suitable for our dataset. Let $(\vx_n, \vy_n, \hat{\vy}_n)$ denote one example that consists of a (source, target, prediction) tuple. PARENT is defined at an instance level as:
\begin{align*}
&PARENT(\vx_n, \vy_n, \hat{\vy}_n) = \\
& \frac{2 \times E_{p}(\vx_n, \vy_n, \hat{\vy}_n) \times E_{r}(\vx_n, \vy_n, \hat{\vy}_n)}{E_{p}(\vx_n, \vy_n, \hat{\vy}_n) + E_{r}(\vx_n, \vy_n, \hat{\vy}_n)}  
\end{align*}
$E_{p}(\vx_n, \vy_n, \hat{\vy}_n)$ is the PARENT precision computed using the prediction, reference, and table (the last of which is not used in BLEU).
$E_{r}(\vx_n, \vy_n, \hat{\vy}_n)$ is the PARENT recall and is computed as:
\begin{align*}
E_{r}(\vx_n, \vy_n, \hat{\vy}_n) = R(\vx_n, \vy_n, \hat{\vy}_n)^{(1-\lambda)} R(\vx_n, \hat{\vy}_n)^{\lambda}
\end{align*}
where $R(\vx_n, \vy_n, \hat{\vy}_n)$ is a recall term that compares the prediction with both the reference and table. $R(\vx_n, \hat{\vy}_n)$ is an extra recall term 
that gives an additional reward if the prediction $\hat{\vy}_n$ contains phrases in the table $\vx_n$ that are not necessarily in the reference ($\lambda$ is a hyperparameter). 

In the original PARENT work, the same table $\vt$ is used for computing the precision and both recall terms. While this makes sense for most existing datasets, it does not take into account the highlighted cells $\vt_{highlight}$ in our task. To incorporate $\vt_{highlight}$, we modify the PARENT metric so that the additional recall term $R(\vx_n, \hat{\vy}_n)$ uses $\vt_{highlight}$ instead of $\vt$ to only give an additional reward for relevant table information. The other recall and the precision term still use $\vt$.

\subsection{Baseline details}


\begin{itemize}
    \item BERT-to-BERT~\citep{rothe2019leveraging} - Uncased model coupling both encoder and decoder as in original paper, with Adam optimizer~\citep{kingma2014adam}. learning rate = 0.05, hidden size = 1024, dropout = 0.1, beam size = 4.
    \item  Pointer Generator~\citep{see2017get} - LSTM with hidden size 300, beam size=8, learning rate = 0.0003, dropout = 0.2, length penalty = 0.0, Adam optimizer~\citep{kingma2014adam}. 
    \item Content planner~\citep{puduppully2019data} - All of the original hyperparameters: content planner: LSTM with hidden size 1x600, realizer LSTM with 2x600, embedding size 600 for both, dropout=0.3, Adagrad optimizer~\citep{duchi2011adaptive}, beam size=5.
\end{itemize}


\end{document}

%% file: math_commands.tex
\usepackage{amsmath,amsfonts,bm}

\def\eqref#1{equation~\ref{#1}}

\def\1{\bm{1}}

\def\vc{{\bm{c}}}
\def\vd{{\bm{d}}}

\def\vf{{\bm{f}}}

\def\vh{{\bm{h}}}

\def\vm{{\bm{m}}}

\def\vs{{\bm{s}}}
\def\vt{{\bm{t}}}

\def\vx{{\bm{x}}}
\def\vy{{\bm{y}}}

\def\mD{{\bm{D}}}

\DeclareMathAlphabet{\mathsfit}{\encodingdefault}{\sfdefault}{m}{sl}
\SetMathAlphabet{\mathsfit}{bold}{\encodingdefault}{\sfdefault}{bx}{n}

